\documentclass[10pt,conference,compsoc]{IEEEtran}
\usepackage{times}
\usepackage{epsfig}
\usepackage{graphicx}
\usepackage[cmex10]{amsmath}
\usepackage{amssymb}
\usepackage{textcomp}
\usepackage{subcaption}
\usepackage{float}

\usepackage[pagebackref=true,breaklinks=true,letterpaper=true,colorlinks,bookmarks=false]{hyperref}
\ifCLASSOPTIONcompsoc
  \usepackage[nocompress]{cite}
\else
  \usepackage{cite}
\fi

\begin{document}
%
\title{Conditional Adversarial Synthesis of 3D Facial Action Units}
\author{
    \IEEEauthorblockN{Zhilei~Liu\IEEEauthorrefmark{1}\IEEEauthorrefmark{2},
    				  Guoxian~Song\IEEEauthorrefmark{2}, 
    				  Jianfei Cai\IEEEauthorrefmark{2},
    				  Tat-Jen Cham\IEEEauthorrefmark{2},
    				  Juyong Zhang\IEEEauthorrefmark{3}}
    				  
    \IEEEauthorblockA{\IEEEauthorrefmark{1}School of Computer Science and Technology, Tianjin University
    \\}
   
    \IEEEauthorblockA{\IEEEauthorrefmark{2}School of Computer Science and Engineering, Nanyang Technological University
    \\}
    
    \IEEEauthorblockA{\IEEEauthorrefmark{3}School of Mathematical Sciences, University of Science and Technology of China
    \\zhileiliu@tju.edu.cn,\{gxsong, ASJFCai, ASTJCham\}@ntu.edu.sg, juyong@ustc.edu.cn}
}

\IEEEtitleabstractindextext{%
\begin{abstract}
Employing deep learning-based approaches for fine-grained facial expression analysis, such as those involving the estimation of Action Unit (AU) intensities, is difficult due to the lack of a large-scale dataset of real faces with sufficiently diverse AU labels for training. In this paper, we consider how AU-level facial image synthesis can be used to substantially augment such a dataset. We propose an AU synthesis framework that combines the well-known 3D Morphable Model (3DMM), which intrinsically disentangles expression parameters from other face attributes, with models that adversarially generate 3DMM expression parameters conditioned on given target AU labels, in contrast to the more conventional approach of generating facial images directly. In this way, we are able to synthesize new combinations of expression parameters and facial images from desired AU labels. Extensive quantitative and qualitative results on the benchmark DISFA dataset demonstrate the effectiveness of our method on 3DMM facial expression parameter synthesis and data augmentation for deep learning-based AU intensity estimation.
\end{abstract}

}

\maketitle

\IEEEdisplaynontitleabstractindextext
\IEEEpeerreviewmaketitle

\section{Introduction}
The Facial Action Coding System (FACS)~\cite{ekman1977FACS} is one of the more comprehensive and objective systems for describing facial expressions, recently receiving increased attention in the field of affective computing and computer vision~\cite{TAC2017Pantic}. It defines a unique set of basic facial muscle actions called Action Units (AUs), each of which has an intensity at a six-point ordinal scale~\cite{walecki2017deep}. Although FACS is helpful for improving reliability, precision, reproducibility and temporal resolution of facial movements, there are still open challenges that impede its widespread use, especially with prevailing state-of-the-art deep learning techniques. First, manual AU annotation is extremely time-consuming and is specialized enough that it requires expert input and cannot be crowdsourced. It takes over 100 hours to train an individual to achieve minimal competency as a FACS coder, and each FACS expert needs approximately one hour to annotate one minute of video~\cite{donato1999classifying}. For this reason, current publicly available facial expression databases with AU labels are limited in size, and cannot be readily applied in state-of-the-art deep learning methods~\cite{TAC2017Pantic}. Secondly, the AU samples in the current databases are highly imbalanced due to the fact that certain AU-intensity combinations rarely appear~\cite{TAC2017Pantic}\cite{walecki2017deep}.

To tackle these challenges, recent deep learning approaches to AU analysis have adopted some heuristical techniques such as iterative balanced batch, simple data augmentation, training on multiple databases, etc.~\cite{gudi2015deep}\cite{TAC2017Pantic}\cite{walecki2017deep}, which will alleviate but not fully solve difficulties with limited AU label data. This provides the motivation for automatic facial expression editing / synthesis, such that desired facial expressions corresponding to given AU labels can be generated, in order to create a large-scale facial expression dataset with accurate and comprehensively diverse AU labels.

\begin{figure}
\centering
\includegraphics[width=0.93\linewidth]{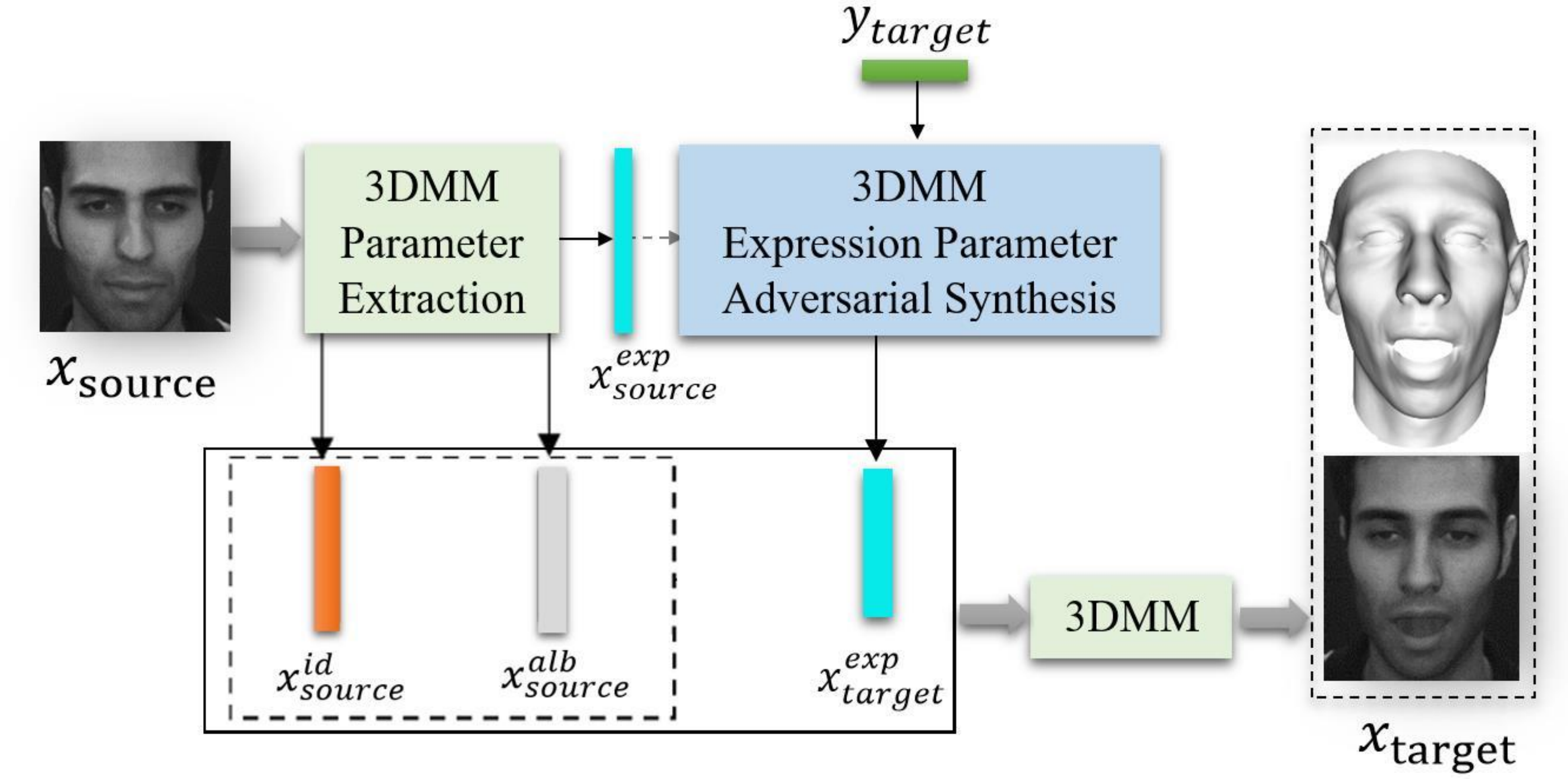}
\caption{Proposed framework for AU synthesis.}
\label{fig:fig_framework}
\end{figure}

An existing geometrical approach to facial image editing / synthesis involves modeling facial key points, texture, shape and other graphical characteristics, and can be used to model facial expressions directly~\cite{lu2017recent}. This approach has been effectively used for facial image synthesis to increase subject diversity~\cite{banerjee2017srefi}. In conjunction with the 3D Morphable Model (3DMM)~\cite{blanz19993DMM}, where a facial image is explicitly parameterized into different facial attributes, facial expressions can be synthesized by directly transferring target facial expression parameters to source facial expression parameters~\cite{thies2015real}\cite{Guo20173DFaceNet}, generating high-resolution facial images while preserving identity and albedo information. However, this direct transfer does not provide the freedom to modify the synthesis based on high-level descriptors such as basic emotions or AUs, which is necessary for dataset diversification.

With the recent development of generative adversarial networks (GANs)~\cite{goodfellow2014generative}, image editing and synthesis have migrated from pixel-level to semantic-level manipulation. Current GAN-based facial editing can be employed on general facial attributes such as facial age~\cite{zhang2017CAAE}, head pose~\cite{zhao2017multi}, etc.~\cite{lu2017recent}, while AU-level expression editing with GAN has also been proposed in~\cite{zhou2017photorealistic}. However, simply relying on GAN models to implicitly disentangle and manipulate discriminative expression descriptors from other facial attributes has met with limited success, with the generated facial images typically of low resolution, with difficulty in depicting differences due to AUs of varying intensities.

In this paper we propose to combine the geometric 3DMM, which intrinsically disentangles expression parameters from other face attributes, with models that adversarially generate 3DMM expression parameters, rather than the conventional approach of generating facial images directly. In particular, these are conditional GANs, with the conditioning on target AU labels. In this way, we are able to generate 3D faces with target expressions specified by different desired AU labels, which can then be rendered to generate the corresponding high-resolution facial images.

Fig.~\ref{fig:fig_framework} shows the diagram of the proposed AU synthesis system. In particular, given a source facial image $x_{source}$, three forms of disentangled 3DMM parameters are first extracted using 3DFaceNet~\cite{Guo20173DFaceNet}: identity $x^{id}_{source}$, albedo $x^{alb}_{source}$, and expression $x^{exp}_{source}$ parameters. Next, target 3DMM expression parameters $x^{exp}_{target}$ are synthesized given any desired target AU label (or combination of labels) $y_{target}$ by using conditional generative adversarial models. Finally, the facial image $x_{target}$ corresponding to the target AU label is synthesized using 3DMM, with preservation of the original identity and albedo parameters.

The contributions of this paper are threefold. First, we propose an AU synthesis framework\footnote{The source code will be publicly released.} that combines a 3DMM with conditional generative adversarial models, which enables high-resolution facial expression synthesis given any desired combination of AU labels. To the best of our knowledge, this has not been done before. Second, with the aid of our newly developed AU synthesis tool, we constructed a large-scale facial expression dataset containing multiple subjects with diverse combinations of AU label categories and intensities\footnote{The constructed facial AU dataset will also be publicly released.}, many of which rarely appear in existing datasets. Third, we conducted extensive experiments on the benchmark dataset to demonstrate the utility of synthesized expression parameters for training data augmentation in deep learning-based AU intensity estimation.

\section{Related Work} 

Our proposed expression synthesis framework is closely related to the existing geometric model based expression editing methods as well as generative model based approaches, since we combine both the 3D geometry model and the generative models.

\textbf{Expression Editing with Geometric Models.} In general, facial expression editing requires to preserve other facial attributes, such as identity, pose, illumination, age, gender, and so on. Given a facial image, geometric models are powerful on facial parameter disentanglement and face reconstruction regarding different facial attributes. Facial expression generation on a new subject using geometric models is often realized by direct expression transfer, which simply replaces the source facial expression parameters with the target ones. 3DMM~\cite{blanz19993DMM} is one of the most well-known 3D geometric models on face modeling, which disentangles a facial image into three types of parameters including face identity, facial expression, and albedo. Aldrian et al.~\cite{aldrian2013inverse} proposed a complete framework for face inverse rendering with 3DMM by decomposing the facial image into geometric and photometric parts. With the help of 3DMM, Thies et al.~\cite{thies2015real} proposed a real-time expression transfer method by computing the difference between the source (neutral expression) and target expressions of one subject in 3DMM expression parameter space, and then adding the difference to the neutral expression of another subject to generate the target expression. Recently, Guo et al.~\cite{Guo20173DFaceNet} presented a coarse-to-fine CNN framework for real-time dense textured 3D face reconstruction with 3DMM facial parameters.

All the above research works are effective on facial expression transfer with high image quality by disentangling expression parameters and replacing them with target ones. However, they cannot realize facial expression editing or synthesis based on high-level semantic descriptions, such as general emotions or precise AU labels. In other words, the mapping between the target emotion or the target AU label and the corresponding expression parameters of the geometry model must exist in the training data. If the mapping does not exist in the training data, direct expression transfer cannot generate the expression of a target emotion or AU label. In contrast, our framework can generate the expression parameters of any AU label and synthesize the corresponding facial image.

\textbf{Expression Editing with Generative Models.} Image editing with generative adversarial models at pixel-level has achieved increasing attention in recent years, and much progress has been made in the field of facial expression editing by applying and altering generative adversarial models. Particularly, Radford et al.~\cite{radford2015DCGAN} proposed a variation of GAN named DCGAN, which can realize facial expression transfer with the help of vector arithmetic on facial images. Huang et al.~\cite{Huang2017DyadGAN} proposed a conditional GAN approach called DyadGAN with two-level optimization to generate contextually valid facial expressions in dyadic human interactions. Ding et al.~\cite{ding2017exprgan} proposed an Expression Generative Adversarial Network (ExprGAN) inspired by CAAE (Conditional Adversarial Autoencoder)~\cite{zhang2017CAAE} and InfoGAN~\cite{chen2016infogan} for photorealistic facial expression editing with controllable expressions with different intensities. However, the generated facial images of all these methods above only focus on facial expression with several general emotions, and the resolutions of the generated images are still not high enough to reflect the local dynamic changes of AUs.

Recently, Zhou et al.~\cite{zhou2017photorealistic} proposed a conditional difference adversarial autoencoder (CDAAE) for facial expression synthesis that considers AU labels. However, the resolution of its generated facial image is only 32$\times$32, and the generated facial images with AU labels are not well quantitatively evaluated. In addition, all these existing generative model based approaches require large training database to implicitly disentangle and manipulate those discriminative expression descriptions apart from other facial attributes over facial images. In contrast, we combine the geometry model of 3DMM, which inherently disentangle expression parameters from other facial attributes, with the generative model by adversarially generating the 3DMM expression parameters conditioned on the target AU label. In this way, we are able to generate high-resolution facial images with the target expression specified by the AU label.

\section{Methodology}

\begin{figure*}[htb]
\begin{subfigure}[b]{.5\textwidth}
  \centering
  \includegraphics[width=.94\linewidth]{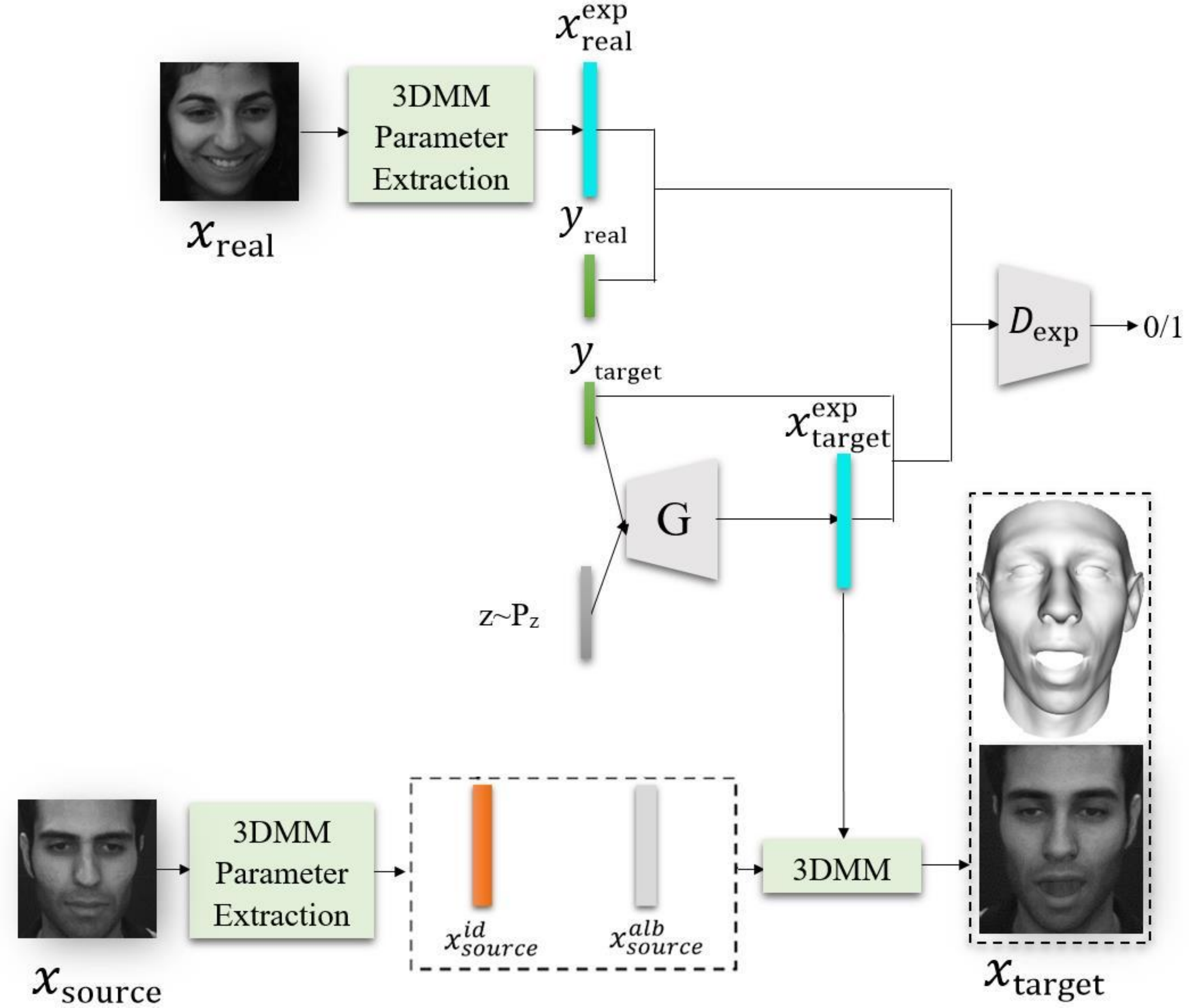}
  \caption{3D facial action unit synthesis with CGAN and 3DMM.}
  \label{fig:fig_CGAN_3DMM}
\end{subfigure}%
\begin{subfigure}[b]{.5\textwidth}
  \centering
  \includegraphics[width=.99\linewidth]{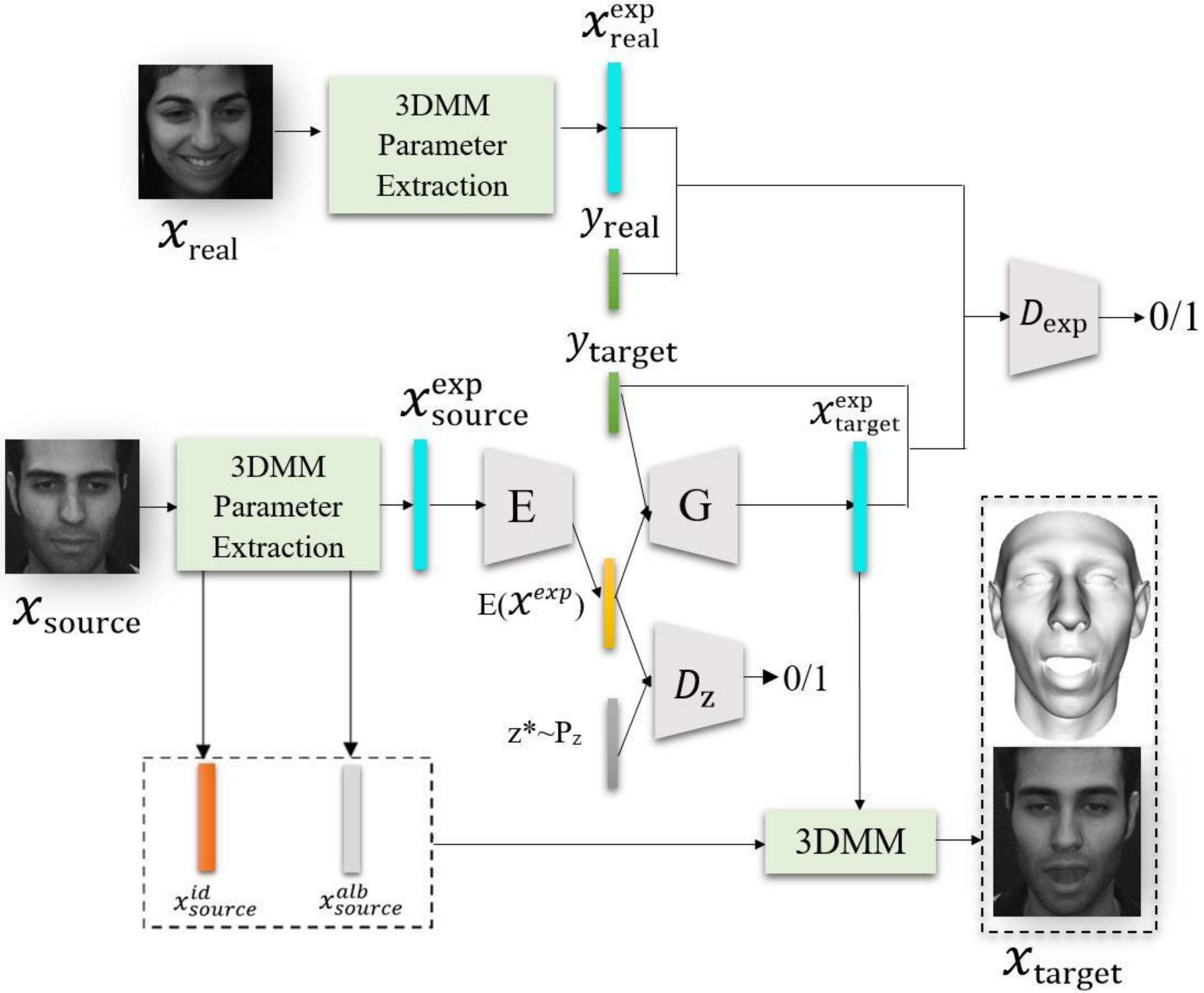}
  \caption{3D facial action unit synthesis with CAAE and 3DMM.}
  \label{fig:fig_CAAE_3DMM}
\end{subfigure}
\caption{Proposed architectures for 3D facial action unit synthesis.}
\label{fig:fig_pipeline}
\end{figure*}

Fig.~\ref{fig:fig_pipeline} shows two proposed architectures for AU synthesis. The basic idea is to avoid directly synthesizing images from target AU labels, but to make use of the 3DMM representation as an intermediary, wherein expression parameters corresponding to the given target AU label are synthesized via conditional adversarial models. In particular,  we consider two kinds of conditional adversarial models for the generation of target 3DMM expression parameters, the first being the Conditional Generative Adversarial Network (CGAN)~\cite{mirza2014CGAN}, with the second being the Conditional Adversarial Autoencoder (CAAE)~\cite{zhang2017CAAE}, as illustrated in Fig.~\ref{fig:fig_CGAN_3DMM} and Fig.~\ref{fig:fig_CAAE_3DMM}, respectively. In the subsequent sections, we first introduce the 3DMM representation and later describe our novel contribution in adapting the two conditional adversarial models with our proposed modified loss functions for AU synthesis via 3DMM expression parameters.

\subsection{3DMM Face Representation}
The 3D morphable model (3DMM)~\cite{blanz19993DMM} is a face representation that encodes 3D geometry and albedo into a lower-dimensional subspace spanned by a well-constructed 3D face dataset.  Specifically, the parametric face model describes 3D face geometry $p$ and albedo $b$ with PCA (principal component analysis) as
\begin{equation}
p = \bar{p}+A^{id}x ^{id} + A^{exp}x ^{exp}
\label{equ:eq_3dmm_exp_representation}
\end{equation}
\begin{equation}
b = \bar{b}+A^{alb}x^{alb}
\label{equ:eq_3dmm_albel_representation}
\end{equation}
where $\bar{p}$ and $\bar{b}$ denote respectively the 3D shape and albedo of the average face, $A^{id}$ and $A^{alb}$ are the principal axes extracted from a set of textured 3D meshes with a neutral expression, $A^{exp}$ represents the principal axes trained on the offsets between the expression meshes and the neutral meshes of individual persons, while $x^{id}$, $x^{exp}$ and $x^{alb}$ are the corresponding coefficient vectors that characterize a specific 3D face model.

\subsection{Generating 3DMM Expression Parameters with CGAN}
Here we use CGAN~\cite{mirza2014CGAN} to generate the expression parameters $x^{exp}_{target}$ of a target 3D face $x_{target}$ with a target AU label $y_{target}$. CGAN is an extension of Generative Adversarial Network (GAN)~\cite{goodfellow2014generative} for conditional distribution generation. As shown in Fig.~\ref{fig:fig_CGAN_3DMM}, it is composed of two networks: a generator network $G$ and a discriminator network $D_{exp}$ that compete in a two-player minimax game. Network $G$ is trained to generate a target expression $x^{exp}_{target}=G(z,y_{target})$ that fools the discriminator $D_{exp}$ into believing that it is sampling from a real expression parameter distribution conditioned on $y_{target}$, where $z$ is random noise following a prior distribution $P_z$. $D_{exp}$ tries to distinguish the distribution of real expression parameters $x^{exp}_{real}$ from that of synthetic parameters $x^{exp}_{target}$, given a target label $y_{target}$.

Denoting the distribution of training data as $P_{data}$, the objective function for $G$ and $D_{exp}$ in the CGAN model can be written as:
\begin{equation}
\begin{split}
\underset{G}{\min}~\underset{D_{exp}}{\max}~&\mathbb{E}_{x^{exp},y\sim P_{data}}[\mathrm{log} D_{exp}(x^{exp},y)] + \\
&\mathbb{E}_{z\sim P_{z},y\sim P_{data}}[\mathrm{log} (1-D_{exp}(G(z,y),y))]
\label{equ:eq_CGAN}
\end{split}
\end{equation}
To avoid significant mesh distortion, we further constrain the range of generated expression parameters by introducing an additional regularization term $L_R$:
\begin{equation}
\begin{split}
\underset{G}{\min}~L_{R} =& \lVert \max (G(z,y)-x^{exp}_{upper},0) \rVert_{L_{1}} + \\
& \lVert  \max (x^{exp}_{lower}-G(z,y),0)  \rVert_{L_{1}}
\label{equ:eq_CGAN_Regulation}
\end{split}
\end{equation}
where $x^{exp}_{upper}$ and $x^{exp}_{lower}$ are the upper and lower limits of the expression parameters.

The overall objective function of the CGAN model for expression parameter generation then becomes
\begin{equation}
\begin{split}
\underset{G}{\min}~\underset{D_{exp}}{\max}~&\beta L_R + \mathbb{E}_{x^{exp},y\sim P_{data}}[\mathrm{log} D_{exp}(x^{exp},y)] + \\
&\mathbb{E}_{z\sim P_{z},y\sim P_{data}}[\mathrm{log} (1-D_{exp}(G(z,y),y))]
\label{equ:eq_CGAN_overall}
\end{split}
\end{equation}
where $\beta$ is a tradeoff parameter.

\subsection{Generating 3DMM Expression Parameters with CAAE}
When CGAN is used to synthesize 3DMM expression parameters, these generated parameters are \emph{independent} of the source facial image. However, it empirically turns out that expression parameters are not fully disentangled from the identity and albedo parameters within the 3DMM, posing a significant problem -- unlike AUs which are universal labels, new expression parameters cannot simply be combined with other existing parameters without risk of producing strange artifacts, because certain expression parameters may in fact be incompatible with the identity and albedo parameters. Thus in this section we adapt another conditional adversarial model, the Conditional Adversarial Autoencoder (CAAE), for generating target 3DMM expression parameters $x^{exp}_{target}$.

CAAE~\cite{zhang2017CAAE} is a conditional extension of the Adversarial Autoencoder (AAE)~\cite{makhzani2015AEE}, which is a probabilistic model consisting of an encoder $E$, a generator network $G$, and a discriminator network $D_{exp}$, as shown in Fig.~\ref{fig:fig_CAAE_3DMM}. In addition to the reconstruction loss, the latent vector $z$ = $E(x^{exp}_{source})$ is regularized by an additional adversarial network with discriminator $D_z$ to impose a prior distribution $P_z$ on $z$. Conceptually we may consider the encoder $E(x^{exp}_{source})$ to be distilling the individualized (i.e. person-dependent) aspects of the source expression parameters $x^{exp}_{source}$, to be combined downstream with the target AU label $y_{target}$ for synthesizing the individualized target expression parameters $x^{exp}_{target}$.

Denoting the sample distribution of the training data as $P_{data}$ and random samples from $P_z$ as $z^*$, we train $E$ and $D_z$ in this adversarial network according to the min-max objective function:
\begin{equation}
\begin{split}
\underset{E}{\min}~\underset{D_z}{\max}~&\mathbb{E}_{z^*\sim P_z}[\mathrm{log} D_z(z^*)] + \\
& \mathbb{E}_{x^{exp}\sim P_{data}}[\mathrm{log} (1-D_z(E(x^{exp})))]
\label{equ:eq_CAAE_Z}
\end{split}
\end{equation}

Additionally, in instances when $y_{target}$ is identically the AU label $y_{source}$ in the source image $x_{source}$ (which is the case during training), we want generator $G$ to reproduce the original 3DMM expression parameters, i.e. such that $G(E(x^{exp}_{source}),y_{target})$ becomes $x^{exp}_{source}$. To encourage this, the reconstruction loss used is:
\begin{equation}
\underset{E,G}{\min}~L_{G} =~\mathcal{L}_{x^{exp},y\sim P_{data}}(x^{exp}, G(E(x^{exp}), y))
\label{equ:eq_CAAE_X}
\end{equation}
where we use $L_1$ loss for $\mathcal{L}(~)$.

As was the case for CGAN,  the discriminator $D_{exp}$ attempts to distinguish the data from the training distribution $P_{data}$ and those sampled from the generated distribution $G(E(x^{exp}))$ given conditional label $y$,  while the generator $G$ is expected to generate a distribution $G(E(x^{exp}), y_{target})$ that confounds $D_{exp}$. Thus we train $G$ and $D_{exp}$ with a similar min-max objective function:
\begin{equation}
\begin{split}
\underset{E,G}{\min}~\underset{D_{exp}}{\max}~& \mathbb{E}_{x^{exp},y\sim P_{data}}[\mathrm{log} D_{exp}(x^{exp},y)] + \\
&\mathbb{E}_{x^{exp},y\sim P_{data}}[\mathrm{log} (1-D_{exp}(G(E(x^{exp}),y),y))]
\label{equ:eq_CAAE_GD}
\end{split}
\end{equation}

An additional term $L_R$, similar to that in~\eqref{equ:eq_CGAN_Regulation},  is also used to regularize the generated 3DMM expression parameters:
\begin{equation}
\begin{split}
\underset{E,G}{\min}~L_{R} = &\lVert \max (G(E(x^{exp}),y)-\alpha^{exp}_{upper}, 0) \rVert_{L_{1}} + \\
&\lVert  \max (\alpha^{exp}_{lower}-G(E(x^{exp}),y), 0)  \rVert_{L_{1}}
\label{equ:eq_CAAE_regulation}
\end{split}
\end{equation}

Finally, the overall objective function of this CAAE model for the expression parameter generation is given by
\begin{equation}
\begin{split}
\underset{E,G}{\min}~&\underset{D_z,D_{exp}}{\max}~\lambda L_{G}+ \beta L_{R}+\\
&\mathbb{E}_{z^*\sim P_z}[\mathrm{log} D_z(z^*)] + \\
&\mathbb{E}_{x^{exp}\sim P_{data}}[\mathrm{log} (1-D_z(E(x^{exp})))]+\\
& \mathbb{E}_{x^{exp},y\sim P_{data}}[\mathrm{log} D_{exp}(x^{exp},y)] + \\
&\mathbb{E}_{x^{exp},y\sim P_{data}}[\mathrm{log} (1-D_{exp}(G(E(x^{exp}),y),y))]
\label{equ:eq_CAAE}
\end{split}
\end{equation}
where $\lambda$ and $\beta$ are tradeoff parameters.

\subsection{Soft Label Processing}
In order to cope with subjective variance in labeling AUs in the training set and the small number of discrete AU intensity levels, a simple but effective soft label processing is introduced by adding a random noise $\delta_i$ to the discrete label $y_i$, which can be expressed as
\begin{equation}
\begin{split}
\bar{y}_i = y_i + \delta_i, ~~~~\delta_i\sim \mathcal{N}(-0.5,0.5)
\label{equ:eq_softlabel}
\end{split}
\end{equation}
where $y_i \in \{0,1,2,3,4,5\} $ is the label of the $i$-th AU with 6 discrete intensity levels, and $\bar{y}_i$ is the corresponding soft label with continuous intensity.

\section{Experiments and Discussions}

\subsection{DISFA Dataset}\label{sec:dataset}

\begin{figure}[htb]
\centering
\includegraphics[width=.95\linewidth]{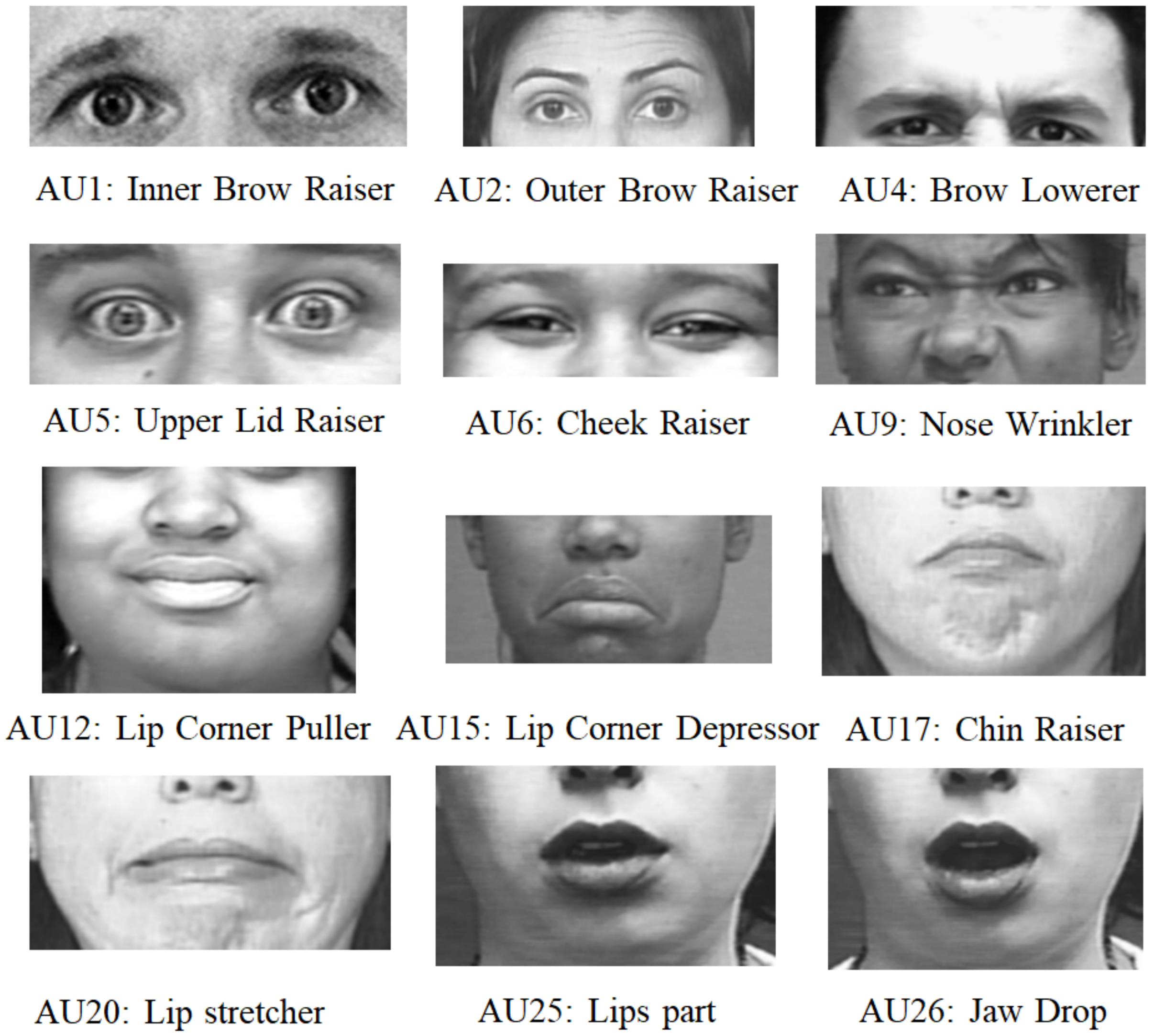}
\caption{The 12 Action Units coded in DISFA database.}
\label{fig:AU}
\end{figure}

There are only two widely studied datasets with AU intensity labels: DISFA~\cite{mavadati2013disfa} and BP4D~\cite{zhang2013high}. However, BP4D only has AU intensity labels for 5 AUs, while DISFA has the intensity labels for 12 AUs. Thus, we only consider DISFA dataset for evaluation. The DISFA dataset consists of 27 facial videos of young adults, which were recorded by a stereo camera while the subjects were viewing video clips, with the intention of capturing spontaneous emotive expressions. Each video frame is manually coded with an intensity for each of the 12 AUs, namely AU1, AU2, AU4, AU5, AU6, AU9, AU12, AU15, AU17, AU20, AU25, and AU26, with intensities ranging from 0 to 5 according to the FACS~\cite{ekman1977FACS} specification, with exemplars shown in Fig.~\ref{fig:AU}. In this work, we used all annotated video frames with successful face registration (233,648 frames) of 25 subjects as the training set for our proposed 3D facial action unit synthesis models, while all annotated video frames (19,380 frames) for the remaining two subjects were used as the test set.

\subsection{Implementation Details}

\begin{figure*}[!htb]
\small
\begin{subfigure}[b]{.16\textwidth}
\centering
  \includegraphics[width=1\linewidth]{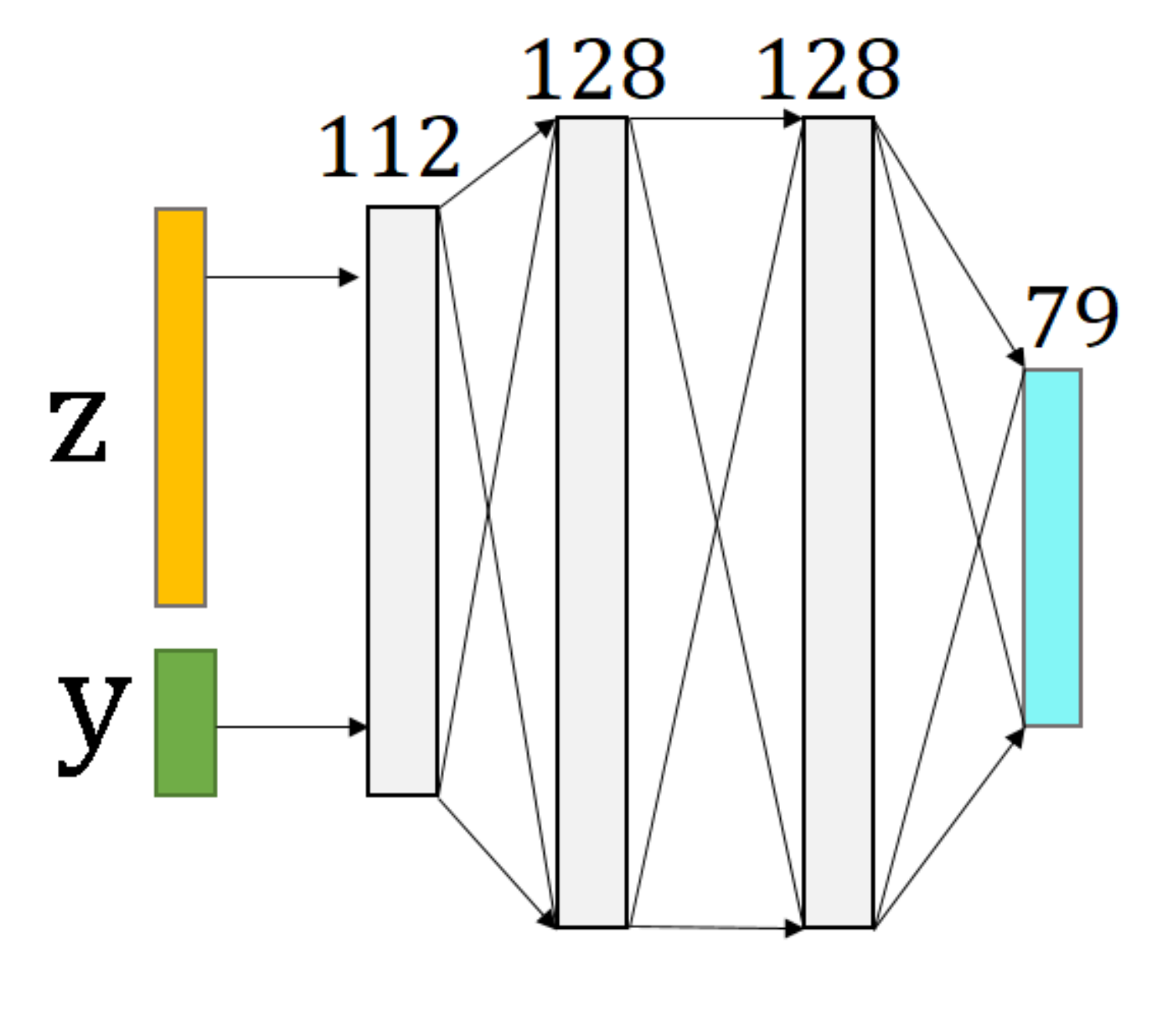}
  \caption{CGAN  $G$}
  \label{fig:fig_CGANG}
\end{subfigure}
\begin{subfigure}[b]{.16\textwidth}
\centering
  \includegraphics[width=1\linewidth]{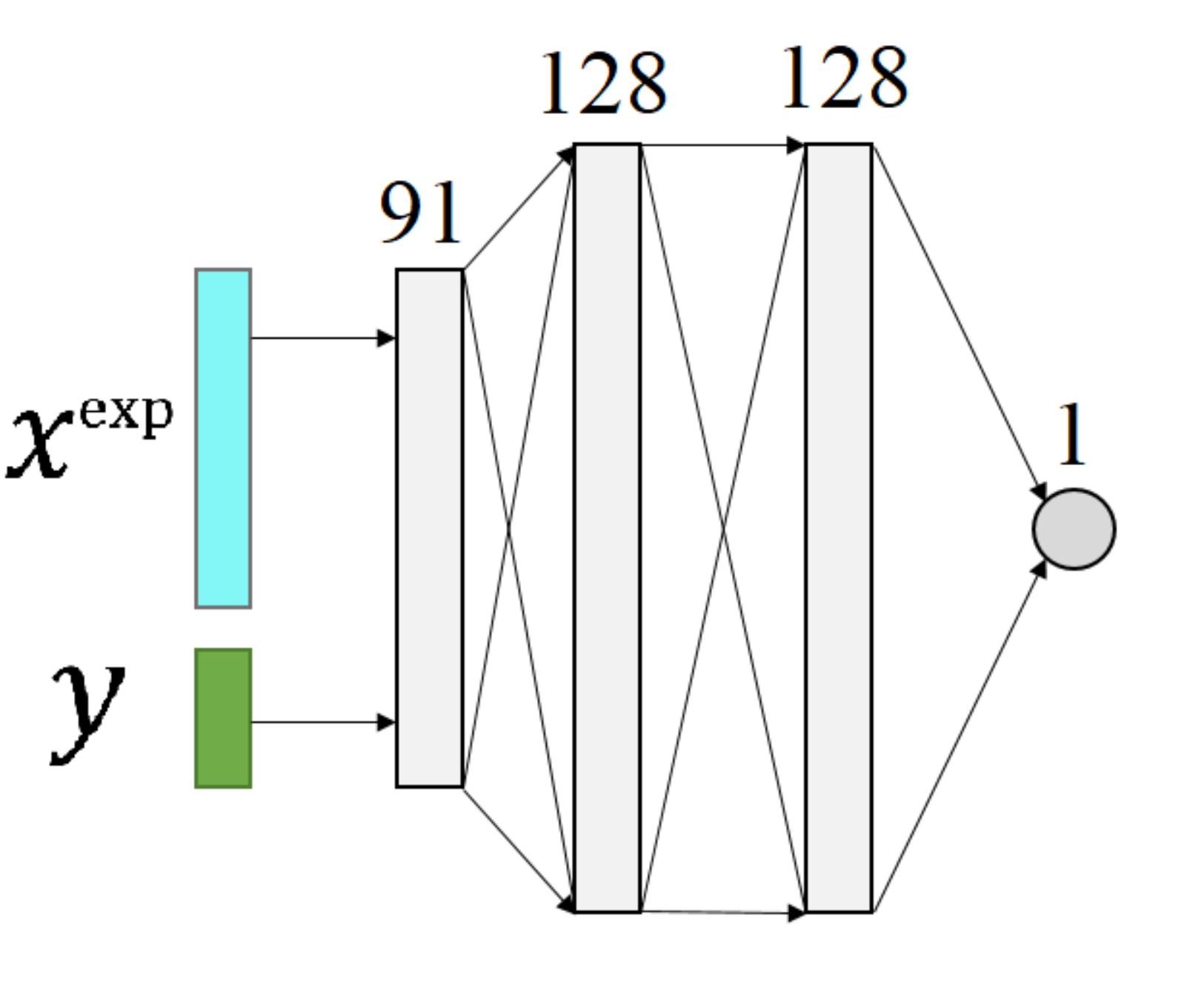}
  \caption{CGAN  $D_{exp}$}
  \label{fig:fig_CGAND}
\end{subfigure}
\begin{subfigure}[b]{.16\textwidth}
\centering
  \includegraphics[width=0.8\linewidth]{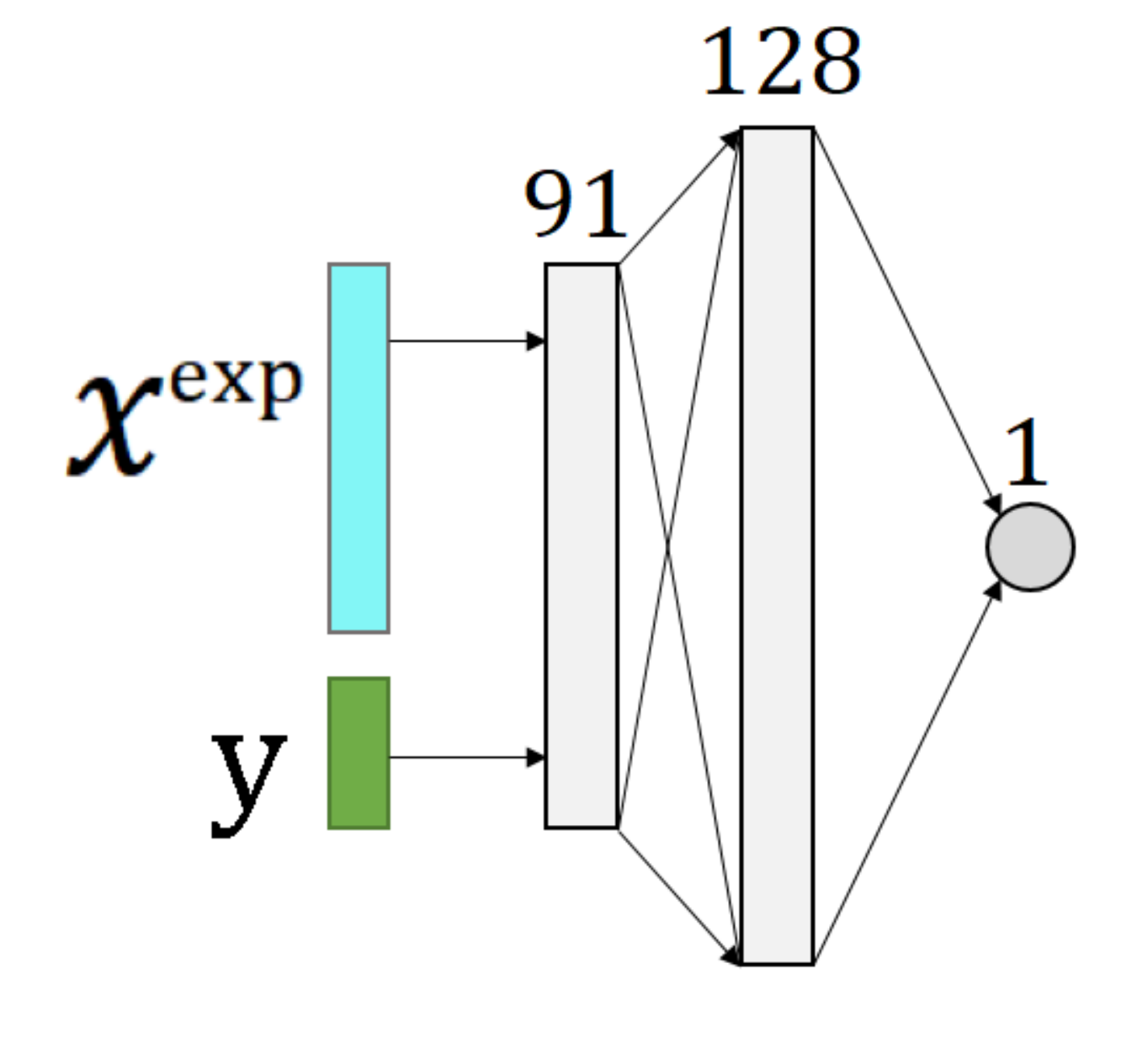}
  \caption{CAAE  $D_{exp}$}
  \label{fig:fig_Dexp}
\end{subfigure}
\begin{subfigure}[b]{.16\textwidth}
\centering
  \includegraphics[width=1\linewidth]{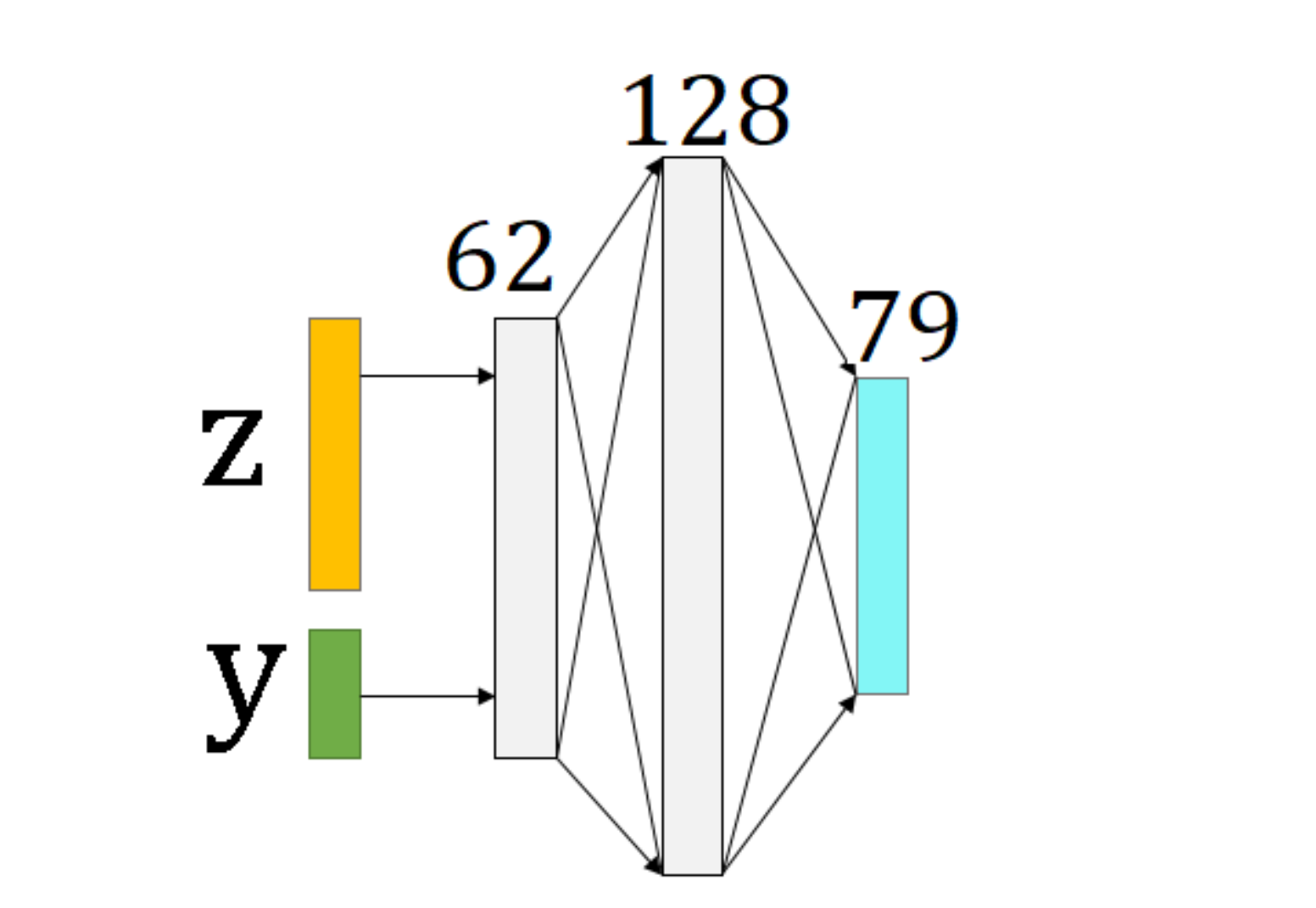}
  \caption{CAAE  $G$}
  \label{fig:fig_Decoder}
\end{subfigure}
\begin{subfigure}[b]{.16\textwidth}
\centering
  \includegraphics[width=0.8\linewidth]{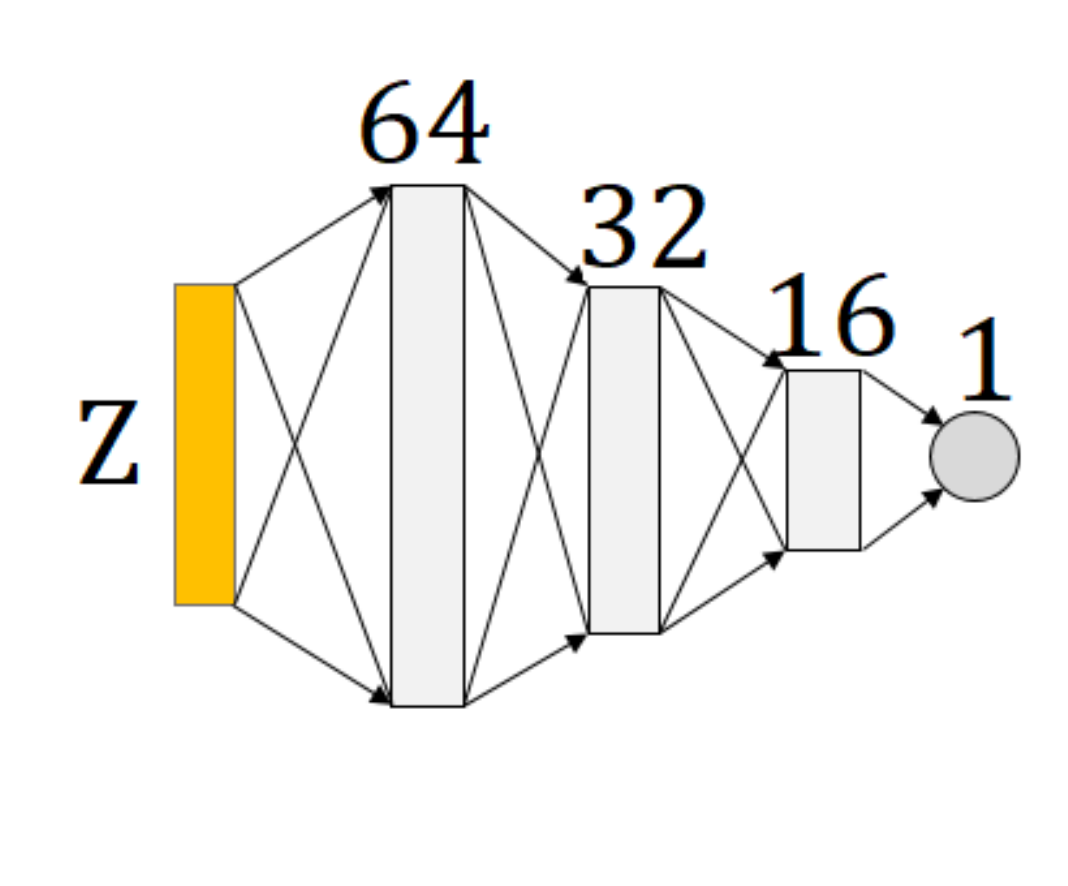}
  \caption{CAAE  $D_{z}$}
  \label{fig:fig_Dz}
\end{subfigure}
\begin{subfigure}[b]{.16\textwidth}
\centering
  \includegraphics[width=0.65\linewidth]{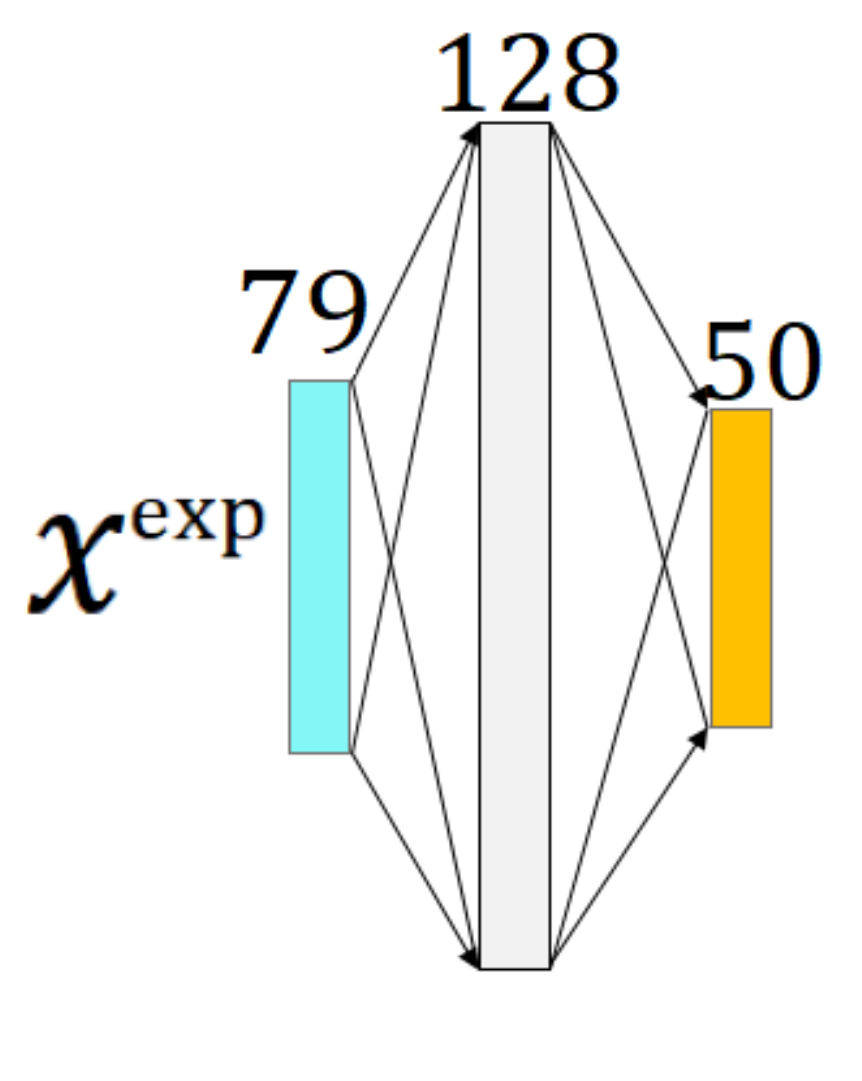}
  \caption{CAAE  $E$}
  \label{fig:fig_Encoder}
\end{subfigure}

\caption{Structural details of sub-networks in the CGAN and CAAE models.}
\label{fig:SubNetStructure}
\end{figure*}

We used 3DFaceNet~\cite{Guo20173DFaceNet} to extract 3DMM coefficients $\{x^{id},~x^{alb},~x^{exp}\}$, where $x^{id}$ and $x^{alb}$ are of 100 dimensions, with the bases from the Basel Face Model (BFM)~\cite{paysan20093d}, while $x^{exp}$ is of 79 dimensions with the bases from FaceWarehouse~\cite{cao2014facewarehouse}.

All extracted 3DMM expression parameters are normalized within $[-1,1]$, and the AU labels with six intensities ranging from 0 to 5 are also linearly scaled to $[-1,1]$. Given a target label vector $y_{target}$ with 12 dimensions denoting the intensity values of the 12 AUs, the CGAN and CAAE models were used to generate the corresponding 3D facial expression parameters $x^{exp}_{target}$. For the CGAN model shown in~Fig.~\ref{fig:fig_CGAN_3DMM}, the structure of its generator $G$  has three fully connected layers as shown in Fig.~\ref{fig:fig_CGANG}, for which the input is a combination of the target label vector of 12 dimensions and the prior $z$ of 100 dimensions, and the output is a 3D facial expression parameter $x^{exp}_{target}$ of 79 dimensions. The structure of the discriminator $D_{exp}$ also has three fully connected layers as shown in Fig.~\ref{fig:fig_CGAND}, where the input is a combination of the 12-dimensional label vector and the 3D expression parameter, while the output is a single-value prediction of the authenticity of the joint distribution of AU labels and expression parameters. The tradeoff parameter $\beta$ in~\eqref{equ:eq_CGAN_overall} is empirically set as 10.

For the CAAE model shown in Fig.~\ref{fig:fig_CAAE_3DMM}, the structures of the expression discriminator $D_{exp}$ and generator $G$ are shown in Figs.~\ref{fig:fig_Dexp}~and~\ref{fig:fig_Decoder} respectively. The encoder $E$ shown in Fig.~\ref{fig:fig_Encoder} disentangles a latent space containing individual-dependent information from the source expression parameters $x^{exp}_{source}$, for which the input is $x^{exp}_{source}$ and the output is a 50-dimensional latent-space vector with an element value range of $[-1, 1]$. The prior discriminator $D_z$, shown in Fig.~\ref{fig:fig_Dz}, is expected to distinguish the output of $E$ from a predefined uniform distribution $P_z$. For all these sub-networks, we use \textit{ReLU}  as the activation function in all hidden layers, while \textit{tanh} and \textit{sigmoid} functions are used for the output layers in CGAN and CAAE. During the training stage, the learning rate is set as $10^{-5}$, and 150k iterations are done with a batch size of 64 using the ADAM~\cite{ADAM} minimizer. The tradeoff parameters $\lambda$ and $\beta$ in~\eqref{equ:eq_CAAE} are empirically set as 100 and 10 respectively.

\subsection{Quantitative Evaluation of Synthetic Outputs}
\textbf{Evaluation of synthesized expression parameters.} The goal here is to determine if the expression parameters synthesized via CGAN and CAAE can be accurately classified by third party AU estimators trained only on expression parameters extracted by 3DMM from real face images. These estimators were support vector regression (SVR) and ordinal support vector regression (OSVR)~\cite{Zhao_2016_CVPR}, with expression parameters as input features and used for estimating AU intensities to 6 intensity levels. The training data for these estimators were expression parameters extracted from random samples comprising 30\% of all non-neutral faces in the training set of section~\ref{sec:dataset}. The test data comprised 3D facial expression parameters generated by CGAN and CAAE using neutral faces in the test set of section~\ref{sec:dataset} as source images and combined with target AU labels from non-neutral faces in the test set.

\begin{table}[!htb]
\begin{center}
\caption{AU intensity estimation using SVR and OSVR with the 3DMM expression parameters as features.}
\label{tab:SVR_OSVR} 
\begin{tabular}{c|cc|cc}
\hline
Classifier & \multicolumn{2}{c|}{SVR}        & \multicolumn{2}{c}{OSVR}        \\ \hline
Metric     & MAE            & MSE            & MAE            & MSE            \\ \hline\hline
Real       & 0.489          & 0.549          & 0.454          & 0.485          \\ \hline
CGAN       & 0.606          & 1.011          & 0.49           & 0.728          \\ \hline
CGAN*      & 0.643          & 1.13           & 0.519          & 0.796          \\ \hline
CAAE       & 0.409          & 0.38           & 0.389          & 0.394          \\ \hline
CAAE*      & \textbf{0.395} & \textbf{0.374} & \textbf{0.386} & \textbf{0.344} \\ \hline
\end{tabular}
\end{center}
\end{table}

Table~\ref{tab:SVR_OSVR} shows the AU intensity estimation results using SVR and OSVR. Two kinds of evaluation metrics are considered here, namely mean absolute error (MAE) and mean squared error (MSE). `Real' refers to the results of performing  AU intensity estimation on the the expression parameters extracted directly from the real images of the test set. In addition, the effectiveness of the soft label processing is also evaluated on the synthetic expression parameters of CGAN and CAAE, with the `*' label indicatiing the use of soft label processing in corresponding rows of the table.

From Table~\ref{tab:SVR_OSVR}, we can observed that: 1) The expression parameters generated by CAAE are more authentic than those by CGAN, giving rise to better AU intensity estimation results. This is because the input to generator $G$ has been suitably disentangled, separating the univeral AU label input from the individualized expression parameters extracted by encoder $E$. 2) The generated expression parameters are comparable or even better than those extracted from the ground-truth images with the same AU labels. This is because in the synthesized expression parameters, only desired AU labels are considered, which avoids the influence of other facial movements appearing in the ground-truth facial images. 3) The soft label processing is effective in CAAE, but not well demonstrated in CGAN. This might be because a noise vector $z$ is already an input of CGAN, which limits the effect of additive noise on $y$. 4) By considering the ordinal information of the AU intensity, AU estimation results of OSVR are generally better than those of SVR.

\subsection{Quantitative Evaluation of Data Augmentation}
Our proposed framework is able to generate facial images with varying combinations of AU labels and intensities. In keeping with our original motivation, this enables us to synthesize an extensive set of labeled imagery with a sufficiently diverse and balanced range of AU labels, that may be used for augmenting real datasets when training AU recognition systems. In our experiments, the label diversity of each subject in the training set is increased by randomly selecting different AU label combinations across different subjects in the training set, which are then fed into the trained 3DMM+CAAE framework, whereupon the system synthesized facial images corresponding to the selected AU label combinations.

For the deep-learning based AU intensity estimator, we adapt ResNet-18~\cite{ResNet} by modifying the last layer to a fully connected layer with 12 AU intensity outputs. To give a fair comparison with the state-of-the-art method, i.e the  CCNN method with deep CNN-CRF structure~\cite{walecki2017deep}, we follow the same training-testing split as in~\cite{walecki2017deep}, with 18 subjects for training and 9 subjects for testing. To demonstrate the impact of data augmentation, four synthesized datasets containing 7K, 9K, 10K and 12K instances are used as augmented data for training the ResNet-based estimator. We use the dlib face detector~\cite{dlib} and map each face image to a size of $224\times224$ using similarity normalization~\cite{normalization}. The network is trained with an initial learning rate of $10^{-4}$, with $10\%$ reduction at each epoch. All experiments are performed on a PC with two NVIDIA GTX 1080 GPUs. 



\begin{table}[htb]
\begin{center}
\caption{Estimation performance with different numbers of synthesized images.}
\label{tab:Augmentation}
\begin{tabular}{|c|c|c|c|c|c|}
\hline
Augmentation Size & 0K & 7K & 9K & 10K & 12K \\ \hline
MAE               &0.630   &0.572    &0.563    &\textbf{0.556}     &0.579     \\ \hline
MSE               &0.844   &0.778    &0.764    &\textbf{0.748}     &0.821     \\ \hline
\end{tabular}
\end{center}
\end{table}

Table~\ref{tab:Augmentation} shows the AU intensity estimation results with different amounts of data augmentation. We observe that by adding 10K synthetic images, we can significantly reduce both MAE and MSE from 0.630 to 0.556 and 0.844 to 0.748 respectively. Further increasing augmentation from 10K to 12K synthetic images leads to worse performance, suggesting that the performance has become saturated by then. 


\begin{figure}[htb]
\centering
\includegraphics[width=.99\linewidth]{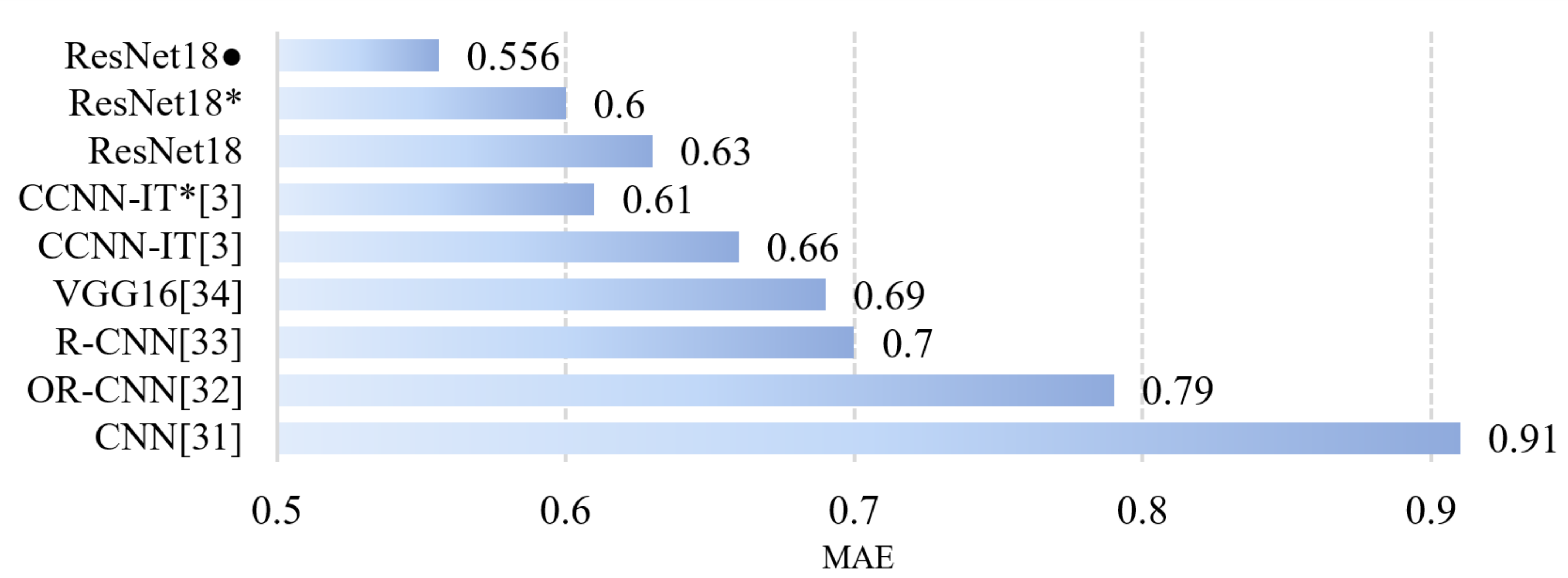}
\caption{Comparison with state-of-the-art deep Models for AU estimation.}
\label{fig:Deep_comparison}
\end{figure}

Fig.~\ref{fig:Deep_comparison} shows the comparison with state-of-the-art deep models, i.e., CNN\cite{CNNAU2015deep}, OR-CNN\cite{ORCNN2016ordinal}, R-CNN\cite{RCNN2016deep},VGG16\cite{VGG2014very}, CCNN-IT and CCNN-IT$\ast$\cite{walecki2017deep} as reported in~\cite{walecki2017deep} on same training and testing data from DISFA, where $\ast$ denotes conventional data augmentation methods like image flipping, random cropping, and so on, and $\bullet$ denotes data augmentation with our synthetic facial images. It can be observed that the AU estimation performance of ResNet augmented with our synthetic facial images outperforms all other state-of-the-art deep models for AU estimation, including ResNet with conventional data augmentation.
\subsection{Visual Results}
In this section, we show visual results of our proposed 3DMM+CAAE framework to illustrate its effectiveness on facial AU synthesis from any desired AU combinations and intensities. Randomly selected neutral faces (with all-zero AU intensity labels) were used as the input to our framework. The visual results are presented as colormaps of 3D mesh deformation between the input neutral faces and the corresponding synthetized faces.

\begin{figure}[htb]
\centering
\includegraphics[width=1\linewidth]{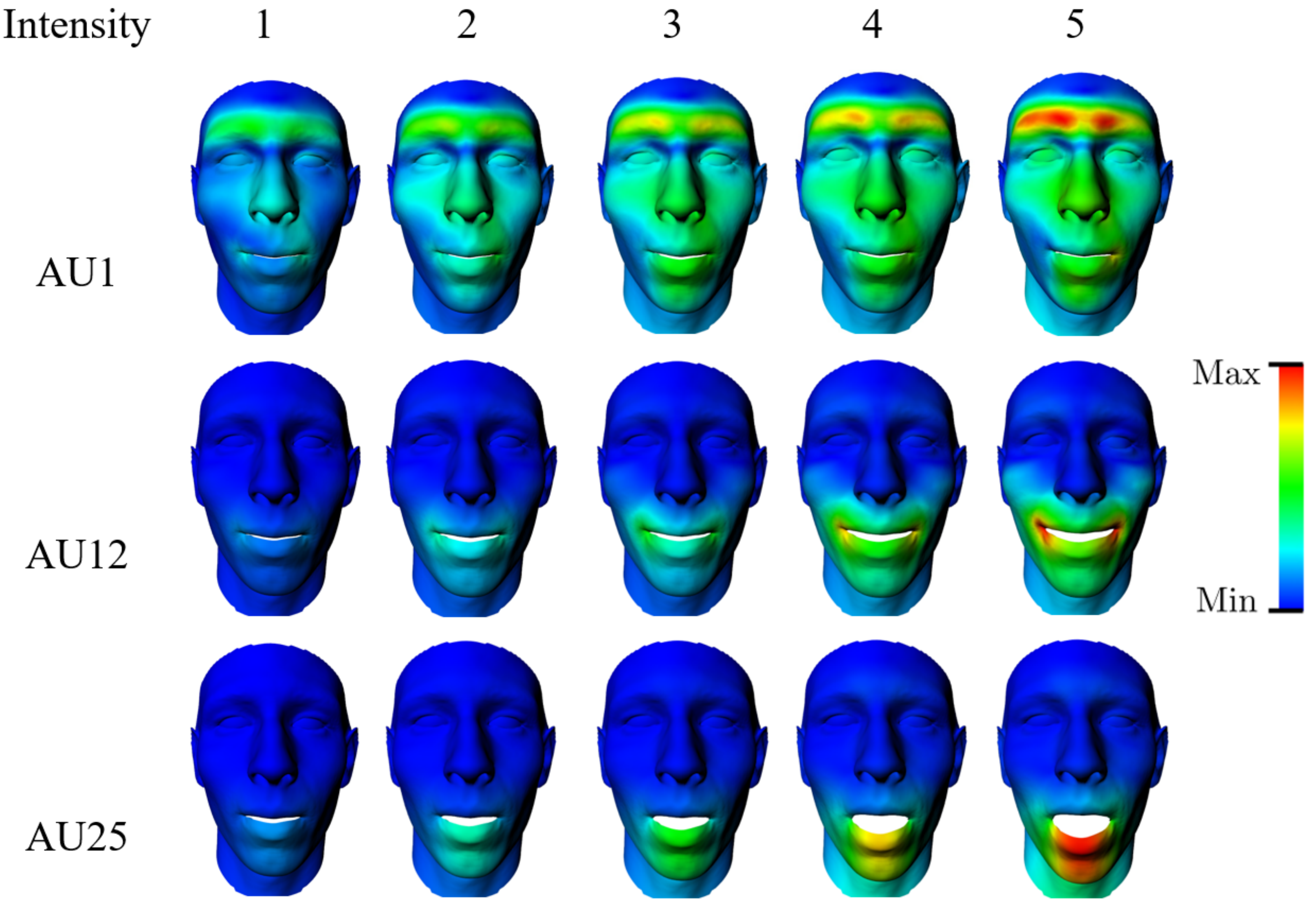}
\caption{Mesh deformation colormaps for AU synthesis with different intensities.}
\label{fig:Syn_intensity}
\end{figure}

Fig.~\ref{fig:Syn_intensity} presents the mesh deformation colormaps for synthetic faces corresponding to three representative AUs over five different intensities, with the 1st, 2nd and 3rd rows respectively showing AU1 (inner brow raiser), AU12 (lip corner puller), and AU25 (lip part) of increasing intensities. Here, brighter colors indicate higher deformation in the respective regions. These visually demonstrates that the synthesized faces correspond to the desired AU labels and the intensities quite well.

\begin{figure}[htb]
\centering
\includegraphics[width=.8\linewidth]{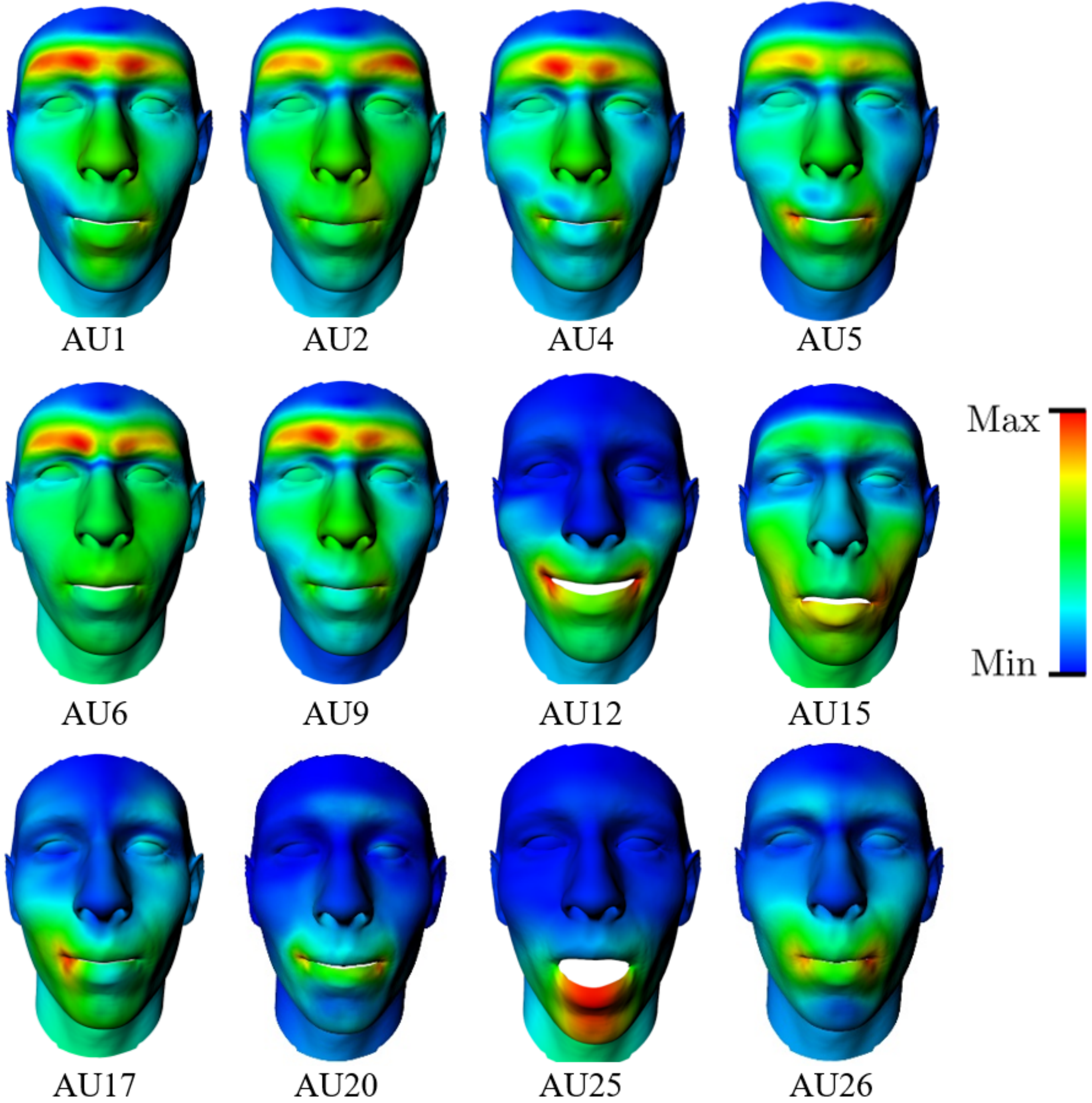}
\caption{Mesh deformation colormaps of 12 AUs.}
\label{fig:Syn_occurrence}
\end{figure}

Fig.~\ref{fig:Syn_occurrence} shows the synthetic outputs corresponding to 12 individual AUs at intensity 5 (setting other AU labels to zero). It can be observed that:
\begin{itemize}
	\item For AU1 (inner brow raiser), AU2 (outer brow raiser) and AU4 (inner brow lowerer) located in the brow region, movement of the eyebrows in the synthetic faces can be clearly seen in the 3D mesh deformation colormaps, verifying the effectiveness of our method. We can also observe that there is co-occurrence between AU1 and AU2, which is reasonable based on typical human facial expressions. It can also been seen from the statistics of DISFA~\cite{wu2017deep} that when AU2 appears, AU1 will also appear with about 90\% probability.
	\item For AU5 (upper lid raiser), AU6 (cheek raiser) and AU9 (nose wrinkler) located in the eye region, their 3D mesh deformation colormaps are not so obvious. This is because all these AUs are described by transient and subtle texture features~\cite{tian2001recognizing}, which cannot be well represented by 3DMM expression parameters. This may be resolved through a more highly detailed geometric model in future work.
	\item For the other AUs corresponding to movement in the mouth and chin regions, clear movement can be seen in the lip corner for AU12 (lip corner puller) and AU15 (lip corner depressor). Subtle changes also appear in the lip and chin regions for AU17 (chin raiser) and AU20 (lip stretcher), which account for minor wrinkling in the synthetic faces. For AU25 (lip part), clear lip changes can be found in the mouth region, which is well synthesized. For AU26 (jaw drop), the synthetic output is not well illustrated. This is because AU26 mostly appears together with AU25 (97\% probability as reported in~\cite{wu2017deep}), and the samples consisting solely of AU26 are very rare and atypical of human facial expression.
\end{itemize}

\begin{figure}[htb]
\centering
\includegraphics[width=1\linewidth]{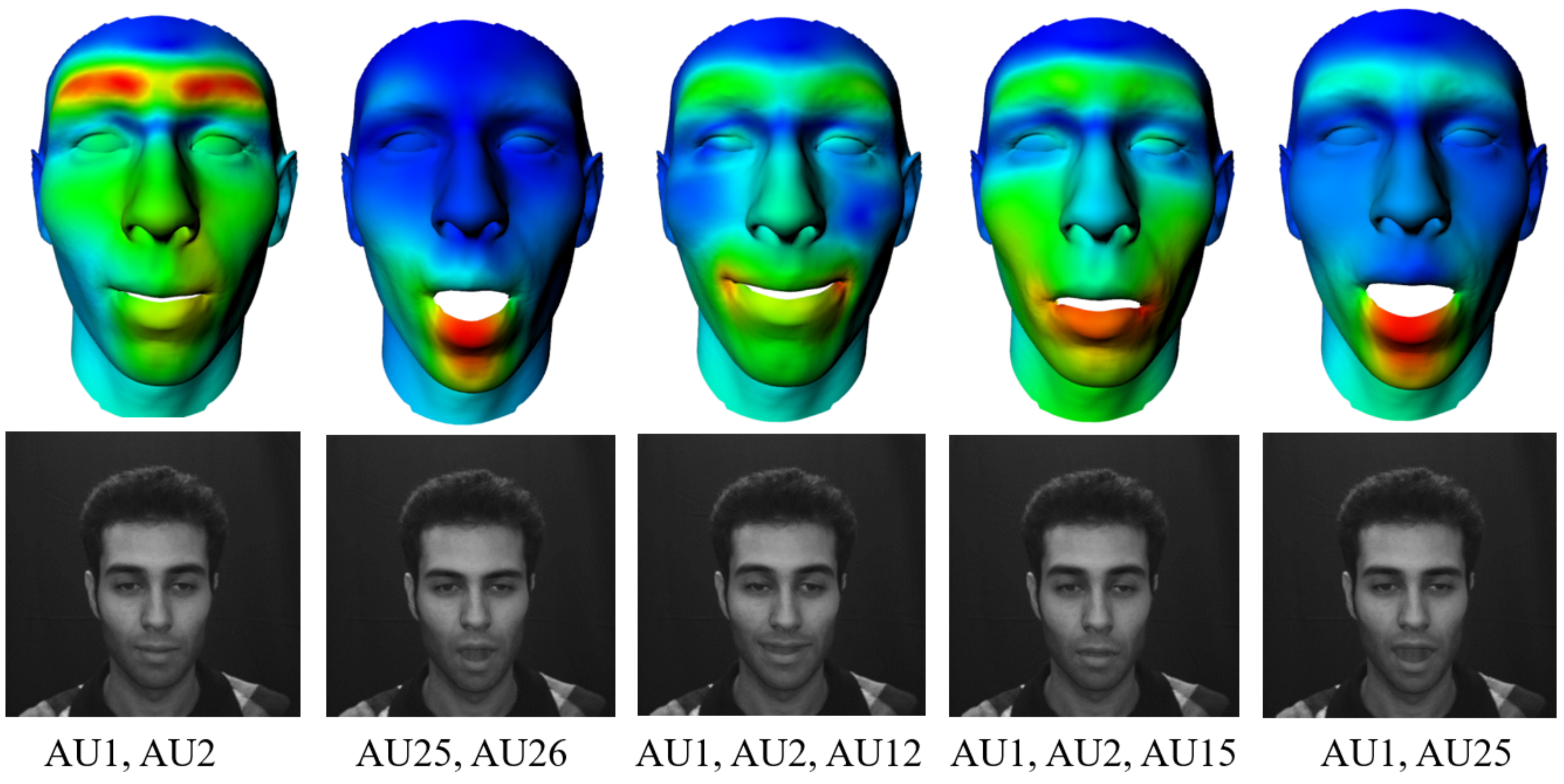}
\caption{Mesh deformation colormaps of AU synthesis with different combinations.}
\label{fig:Syn_combination}
\end{figure} 

Fig.~\ref{fig:Syn_combination} shows the synthetic output of some common AU combinations, from which we can observe that:
\begin{itemize}
	\item Since AU1 and AU2 mostly appear together, the samples with both AU1 and AU2 were well synthesized;
	\item When AU26 is combined with AU25, clear movement can be seen in the lip and jaw regions, which validates our previous comments on AU26;
	\item By varying AU combinations, we can generate different facial expressions as desired based on expression coding principles~\cite{TAC2017Pantic}. The combinations of (AU1, AU2 and AU12), (AU1, AU2 and AU15) and (AU1, AU25)  were successful in creating expressions of happiness, sadness and surprise respectively.
\end{itemize}

\section{Conclusions}
In this paper, we presented a framework for 3D facial action unit synthesis, which combines the advantages of both 3DMM and conditional generative adversarial models. Given a facial image and any desired AU label combinations and intensities, our framework can generate a high-resolution 3D facial image with corresponding facial expressions, while preserving the original identity and albedo information. Extensive quantitative and qualitative results demonstrate the effectiveness of our method on 3DMM facial expression parameter synthesis and data augmentation for deep learning based AU intensity estimation.

Our framework is only the first attempt on combining a 3D geometric model with generative adversarial models for AU synthesis, and there are still areas that need further improvement. One issue is that the 3DMM, being a low-dimensional representation, is unable to represent fine geometric details, which limits the ability of our framework to synthesize subtle AUs. A future implementation with a more complex geometric model may lead to improvement in results.

{\small
\bibliography{3DAU_bib}
}

\end{document}